\documentclass{article}
\usepackage{ijcai20}

\usepackage{times}
\usepackage{soul}
\usepackage{url}
\usepackage{color,xcolor}
\usepackage{graphicx}
\usepackage[hidelinks]{hyperref}
\usepackage[utf8]{inputenc}
\usepackage[small]{caption}
\usepackage{graphicx}
\usepackage{amsmath}
\usepackage{amsthm}
\usepackage{booktabs}
\usepackage{algorithm}
\usepackage{algorithmic}
\usepackage{amssymb}
\usepackage{subfigure}
\title{Multi-hop Reading Comprehension across Documents with Path-based Graph Convolutional Network}

\author{
    Zeyun Tang \and Yongliang Shen \and Xinyin Ma \and Wei Xu \and Jiale Yu \And Weiming Lu\thanks{corresponding author} \\
    \affiliations
    College of Computer Science and Technology, Zhejiang University, China \\
    \emails
    luwm@zju.edu.cn
}

\begin{document}

\maketitle

\begin{abstract}

  Multi-hop reading comprehension across multiple documents attracts much attention recently. In this paper, we propose a novel approach to tackle this multi-hop reading comprehension problem. Inspired by human reasoning processing, we construct a path-based reasoning graph from supporting documents. This graph can combine both the idea of the graph-based and path-based approaches, so it is better for multi-hop reasoning. Meanwhile, we propose Gated-RGCN to accumulate evidence on the path-based reasoning graph, which contains a new question-aware gating mechanism to regulate the usefulness of information propagating across documents and add question information during reasoning. We evaluate our approach on WikiHop dataset, and our approach achieves state-of-the-art accuracy against previously published approaches. Especially, our ensemble model surpasses human performance by 4.2\%.

\end{abstract}

\section{Introduction}

Machine reading comprehension has been a popular topic in the past years, and a variety of models have been proposed to address this problem, such as \textsc{BiDAF}~\cite{Seo2016BidirectionalAF}, Reinforced mnemonic reader~\cite{Hu2017ReinforcedMR}, and ReasoNet~\cite{shen2017reasonet}.
However, most existing works focus on finding evidence and answer in a single document.

In fact, in reality, many questions can only be answered after reasoning across multiple documents. Table~\ref{tab:example} shows a multi-choice style reading comprehension example, which is from \textsc{WikiHop} dataset~\cite{Welbl2018ConstructingDF}. In the example, we can only answer the question `\textit{what is the place of death of alexander john ellis?}' after extracting and integrating the facts `\textit{Alexander John Ellis is buried in Kensal Green Cemetery}' and `\textit{Kensal Green Cemetery is located in Kensington}' from multiple documents, which is a more challenging task.
\begin{table}[htb]
	\centering
	\begin{tabular}{p{8cm}}
		\toprule
		\textbf{Question}: place of death, \textcolor{magenta}{alexander john ellis}, ? \\
		\midrule
		\textbf{Support doc1}: \textcolor{magenta}{Alexander John Ellis}, was an English mathematician ... is buried in \textcolor{orange}{Kensal Green Cemetery}. \\
		\textbf{Support doc2}: The areas of College Park and \textcolor{orange}{Kensal Green Cemetery} are located in the London boroughs of Hammersmith \& Fulham and \textcolor{violet}{Kensington} \& Chelsea, respectively. \\
		......\\
		\midrule
		\textbf{Candidates}: college park, france, \textcolor{violet}{Kensington}, London \\
		\midrule
		\textbf{Answer}: \textcolor{violet}{Kensington} \\
		\bottomrule
	\end{tabular}
	\caption{An example of multi-hop reading comprehension across documents.}\label{tab:example}
\end{table}

The main challenge is that the evidence is distributed in different documents and there is a lot of noise in the documents. We need to extract this evidence from multiple documents, but it is difficult to capture their dependencies for reasoning. Many works used graph convolution networks(GCNs) to deal with this problem, such as Entity-GCN~\cite{de2019question}, BAG~\cite{Cao2019BAGBA} and HDE~\cite{Tu2019MultihopRC}. They transform documents into an entity graph, and then import the entity graph into graph convolution networks(GCNs) to simulate the process of multi-hop reasoning.

However, these GCN-based approaches have some disadvantages. Firstly, they generated the entities only from the question and candidate answers, lacking much key information for multi-hop reasoning. For example, as the example in Table~\ref{tab:example}, the entity `\textit{Kensal Green Cemetery}' is an important clue to answer the question, but the above approaches ignored this information. Secondly, the traditional GCNs only update the central node based on the aggregated information of adjacent nodes and use this to simulate the process of reasoning. But the question information is not fully utilized and there is a lot of irrelevant information during information propagating across documents in the multi-hop reasoning.

In this paper, we propose a novel approach to solve the above problem. We introduce a path-based reasoning graph for multiple documents. Compared to traditional graphs, the path-based reasoning graph contains multiple reasoning paths from questions to candidate answers, combining both the idea of the GCN-based and path-based approaches. Thus, we construct a path-based reasoning graph by extracting reasoning paths(e.g., \textit{Alexander John Ellis $\rightarrow$ Kensal Green Cemetery $\rightarrow$ Kensington}) from supporting documents and then adding reasoning nodes (e.g., \textit{Kensal Green Cemetery}) in these paths to the entity graph. And then, we apply a Gated-RGCN to learn the representation of nodes. Compared to GCNs, Gated-RGCN utilizes attention and question-aware gating mechanism to regulate the usefulness of information propagating across documents and add question information during reasoning, which is closer to human reasoning processes.

Our contributions can be summarized as follows:
\begin{itemize}
    \item We propose a path-based reasoning graph, which introduces information about reasoning paths into the graph;
    \item We propose Gated-RGCN to optimize the convolution formula of RGCN, which is more suitable for multi-hop reading comprehension;
    \item We evaluated our approach on \textsc{WikiHop} dataset~\cite{Welbl2018ConstructingDF}, and our approach achieves new state-of-the-art accuracy. Especially, our ensemble model surpasses the human performance by $4.2\%$.
\end{itemize}

\section{Related Work}
Recently, there are several categories of approaches that have been proposed to tackle the problem of multi-hop reading comprehension across documents, including GCN-based approaches (Entity-GCN~\cite{de2019question}, BAG~\cite{Cao2019BAGBA}, HDE~\cite{Tu2019MultihopRC}, MHQA-GRN~\cite{Song2018ExploringGP}, DFGN~\cite{Qiu2019DynamicallyFG}), memory based approaches (Coref-GRU~\cite{Dhingra2018NeuralMF}, EPAr~\cite{Jiang2019ExplorePA}), path based approaches (PathNet~\cite{DBLP:conf/acl/KunduKSC19}), and attention based approaches (CFC~\cite{Zhong2019CoarsegrainFC}, DynSAN~\cite{Zhuang2019TokenlevelDS}).

GCN-based approaches organize supporting documents into a graph, and then employ Graph Neural Networks based message passing algorithms to perform multi-step reasoning. For example, Entity-GCN~\cite{de2019question} constructed an entity graph from supporting documents, where nodes are mentions of subject entity and candidates, and edges are relations between mentions. BAG~\cite{Cao2019BAGBA} applied bi-directional attention between the entity graph and the query after GCN reasoning over the entity graph. HDE~\cite{Tu2019MultihopRC} constructed a heterogeneous graph where nodes correspond to candidates, documents, and entities. MHQA-GRN~\cite{Song2018ExploringGP} constructed a graph where each node is either an entity mention or a pronoun representing an entity, and edges fall into three types:  same-typed, window-typed and coreference-typed edge.
DFGN~\cite{Qiu2019DynamicallyFG} proposed a dynamic fusion reasoning block based on graph neural networks. Our work proposes Gated-RGCN to optimize the graph convolution operation, it is better for regulating the usefulness of information propagating across documents and add question information during reasoning.

Memory-based approaches try to aggregate evidences for each entity from multiple documents through a memory network. For example, Coref-GRU~\cite{Dhingra2018NeuralMF} aggregated information from multiple mentions of the same entity by incorporating coreference in the GRU layers. EPAr~\cite{Jiang2019ExplorePA} used a hierarchical memory network to construct a `reasoning tree', which contains a set of root-to-leaf reasoning chains, and then merged evidences from all chains to make the final prediction.

PathNet~\cite{DBLP:conf/acl/KunduKSC19} proposed a typical path-based approach for multi-hop reading comprehension. It extracted paths from documents for each candidate given a question, and then predicted the answer by scoring these paths. Our work introduces the idea of path-based approach on GCN-based approach which is better for multi-hop reasoning.

CFC~\cite{Zhong2019CoarsegrainFC} and DynSAN~\cite{Zhuang2019TokenlevelDS} are two typical attention-based approaches. CFC applied co-attention and self-attention to learn query aware node representations of candidates, documents and entities. While DynSAN proposed a dynamic self-attention architecture to determine what tokens are important for constructing intra-passage or cross-passage token level semantic representations. In our work, we employ an attention mechanism between graphs and the question at each layer of Gated-RGCN.

Meanwhile, in order to promote the research on multi-hop QA, several datasets have been designed, including WikiHop~\cite{Welbl2018ConstructingDF}, OpenBookQA~\cite{Mihaylov2018CanAS}, NarrativeQA~\cite{Kocisk2017TheNR}, MultiRC~\cite{Khashabi2018LookingBT} and HotpotQA~\cite{Yang2018HotpotQAAD}. For example, WikiHop is a multi-choice style reading comprehension data set, where the task is to select the correct object entity from candidates when given a query $\langle s,r,? \rangle$ and a set of supporting documents. While OpenBookQA focuses on the multi-hop QA which needs a corpus of provided science facts (open book) with external broad common knowledge.

In addition, \textit{knowledge completion over knowledge graph (KG)} and \textit{KG based query answering} are also related to our task, since they both need multi-hop reasoning, i.e., finding the reasoning path between two entities in KG. For example, MINERVA~\cite{Das2017GoFA} formulates the multi-hop reasoning as a sequential decision problem, and uses the REINFORCE algorithm~\cite{Williams1992SimpleSG} to train an end-to-end model for multi-hop KG query answering. Meta-KGR~\cite{Lv2019AdaptingMK} also used the reinforcement learning method to learn a relation-specific multi-hop reasoning agent to search for reasoning paths and target entities. They further used meta-learning to perform multi-hop reasoning over few-shot relations of knowledge graphs.

\section{Approach}
In this section, we first formulate the task of multi-hop reading comprehension across documents, and then elaborate our approach in detail.

\subsection{Task Formulation}
\begin{figure*}[!htp]
	\centering
	\includegraphics[width=0.9\textwidth]{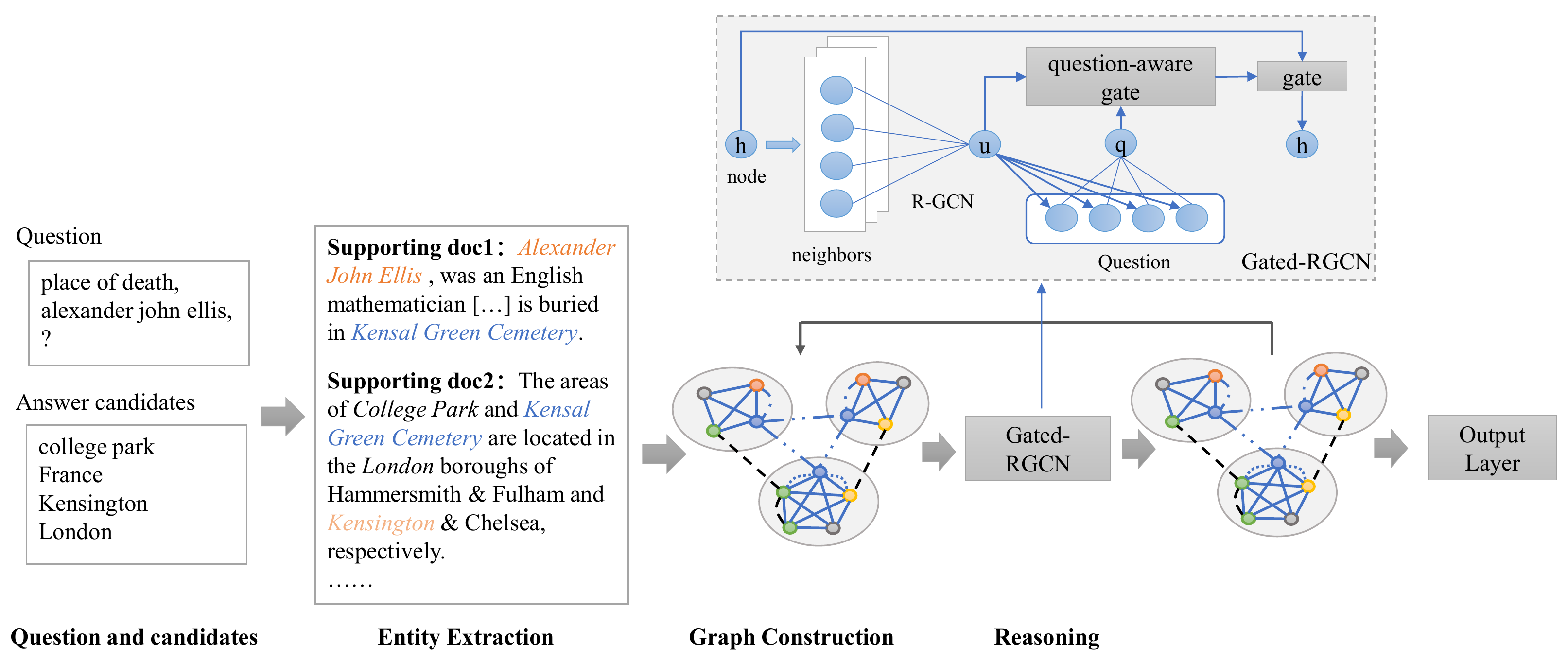}
	\caption{Overview of our approach}\label{fig:arch}
\end{figure*}
The task of multi-hop reading comprehension across documents can be formally defined as: given a question $q=(q_1,q_2,...,q_M)$ and a set of supporting documents $S_q$, the task is to find the correct answer $a$ from a set of answer candidates $C_q=(c_1, c_2, ...,c_N)$, where $M$ is the number of words in the question $q$ and $N$ is the number of candidates in $C_q$.

In the \textsc{WikiHop} dataset~\cite{Welbl2018ConstructingDF}, the question $q$ is given in the form of a tuple $\langle s, r, ? \rangle$, where $s$ represents the subject entity, and $r$ represents the relation between $s$ and the unknown tail entity.
In our example, $q=\langle alexander\ john\ ellis, place\ of\ death, ? \rangle$ means \textit{where did alexander john ellis die}, and the answer candidates is $C_q=(college\ park, france, Kensington, London)$. When given the supporting documents, e.g., \textit{supporting doc1, doc2} in Table~\ref{tab:example}, we should identify the correct answer $a = Kensington$ from the candidates by reasoning across these documents.

As shown in Figure~\ref{fig:arch}, our approach mainly consists of three components, including graph construction, reasoning with Gated-RGCN, and output layer. In the following sections, we will elaborate on each component in detail.

\subsection{Graph Construction}
\label{sec:gc}
We construct an entity graph based on the Entity-GCN~\cite{de2019question}, which extracts all mentions of entities in $C_q \cup \{s\}$ in $S_q$ as nodes in the graph.
Besides, inspired by the human reasoning processing, \textit{reasoning paths} from the subject entity in question to the candidates could be helpful for reasoning across documents, so we add \textit{reasoning entities} in the paths into our entity graph. In our example, the path \textit{alexander john ellis} $\rightarrow$ \textit{Kensal Green Cemetery} $\rightarrow$ \textit{Kensington} from documents indicate that the candidate \textit{Kensington} may be the correct answer for the question $\langle$\textit{alexander john ellis, place of death, ?} $\rangle$. Thus, we treat \textit{Kensal Green Cemetery} as a reasoning entity, and add it into the entity graph.

Formally, for a given question $q=\langle s,r,? \rangle$, we would like to extract paths from $s$ to $c_i \in C_q$ from $S_q$, e.g., $p_{i}=s \rightarrow e_1 \rightarrow {e_2} \rightarrow ... \rightarrow e_{l} \rightarrow c_{i}$, where $e_i$ is a reasoning entity.
In order to find a path, we first find a document $d_1$ which contains the mention $m_s$ of the subject entity $s$ in $S_q$, and then find all the named entities and noun phrases that appear in the same sentence with $m_s$. In our example, we find \textit{Kensal Green Cemetery} and \textit{alexander john ellis} appear in the same sentence in \textit{supporting doc1}, so we extract \textit{Kensal Green Cemetery} as one of reasoning entities. Then, we find another document $d_2$ which contains any of the reasoning entities. In our example, \textit{supporting doc2} contains the reasoning entity \textit{Kensal Green Cemetery}. Finally, we check whether the reasoning entity appears with one of the candidates in the same sentence. If so, we would add the path to the entity graph. For example, \textit{Kensal Green Cemetery} and \textit{Kensington} appear in the same sentence in \textit{supporting doc2}. Therefore, the path \textit{alexander john ellis} $\rightarrow$ \textit{Kensal Green Cemetery} $\rightarrow$\textit{Kensington} can be added to the entity graph.

\begin{figure}[!htp]
	\centering
	\includegraphics[width=0.8\linewidth]{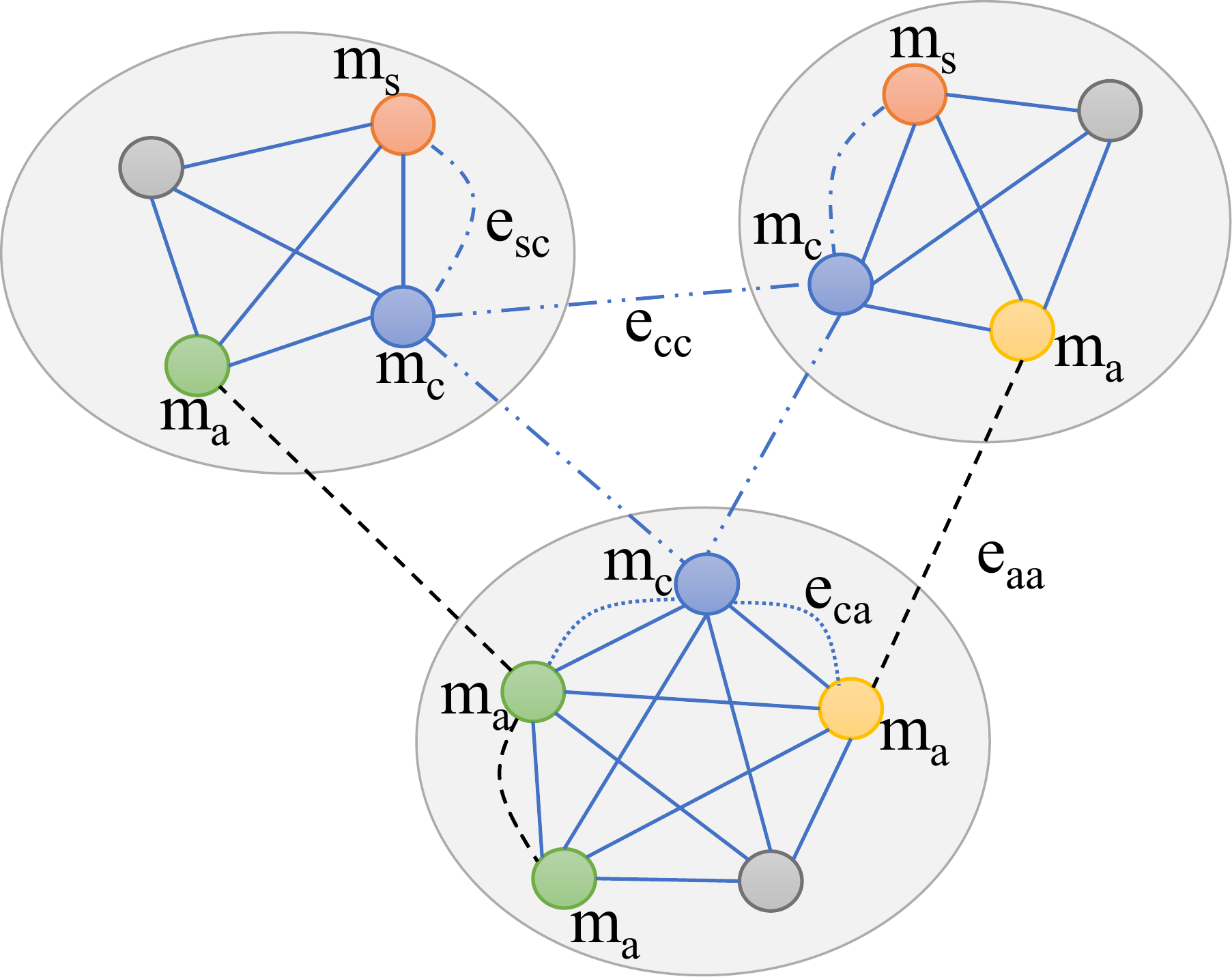}
	\caption{The entity graph for supporting documents (ellipses), where nodes are mentions of the subject entity ($m_s$), reasoning entities ($m_c$), and candidates ($m_a$). Nodes with same color indicate they refer to the same entity. Meanwhile, nodes are connected by six types of edges, which are elaborated in Section~\ref{sec:gc}.} \label{fig:egraph}
\end{figure}

Since each entity in different documents has different contexts, so we use mentions of the subject entity, reasoning entities, and candidate answers as nodes in the entity graph. In our example, \textit{Kensal Green Cemetery} appears in two different sentences, so we need to add nodes for different positions Figure~\ref{fig:egraph} shows an example of an entity graph, where $m_s$, $m_c$ and $m_a$ are mentions of the subject entity $s$, reasoning entities $C_c$, and candidate answers $C_q$ respectively.

Then, we define the following types of edges between pairs of nodes to encode various structural information in the entity graph.
\begin{enumerate}
	\item an edge between a subject node and a reasoning node if they appear in the same sentence in a document, e.g., $e_{sc}$ in Figure~\ref{fig:egraph}.
	\item an edge between two nodes if they are reasoning nodes and are adjacent nodes on the same path, e.g., $e_{cc}$ in Figure~\ref{fig:egraph}.
	\item an edge between a reasoning node and a candidate node if they appear in the same sentence in a document, e.g., $e_{ca}$ in Figure~\ref{fig:egraph}.
	\item an edge between two nodes if they are mentions of the same candidate, e.g., $e_{aa}$ in Figure~\ref{fig:egraph}.
	\item an edge between two nodes if they appear in the same document.
	\item nodes that do not meet previous conditions are connected.
\end{enumerate}

\subsection{Reasoning with Gated-RGCN}
We first use pretrained word embeddings GLoVe~\cite{Pennington2014GloveGV} to represent each node in the entity graph, and then use ELMo~\cite{peters2018deep} to model contextual information for each node in different documents. These two vectors are concatenated, and then encoded through 1-layer linear network. Thus, the features for all nodes can be denoted as $f_n \in \mathbb{R}^{T \times d}$, where $T$ is the number of nodes in the graph, and $d$ is the dimension of the node feature.

After graph initialization, we employ a Gated Relational Graph Convolutional Network (Gated-RGCN) to realize multi-hop reasoning. First, we use R-GCN to aggregate messages from its direct neighbors. Specifically, at $l$th layer, the aggregated message $z_i^l$ for node $i$ can be obtained via
$$
z_i^l = \sum_{j\in {\mathcal{N}_i}}\sum_{r\in R_{ij}} \frac{1}{|\mathcal{N}_i|}W_r^l h_j^l
$$
where $\mathcal{N}_i$ is the neighbors of node $i$, $R_{ij}$ is the set of relations between $i$ and $j$, $W_r^l \in \mathbb{R}^{d \times d}$ is a relation-specific weight matrix, $|\cdot|$ indicates the size of $\mathcal{N}_i$, and $h_j^l$ is the hidden state of node $j$ at $l$th layer. Then, the update message $u_i^l$ for node $i$ can be obtained by combining the aggregated message with its original node information:
$$
u_i^l = W_0^l h_i^l + z_i^l
$$
where $W_0^l \in \mathbb{R}^{d \times d}$ is a general weight.

In Entity-GCN~\cite{de2019question}, HDE~\cite{Tu2019MultihopRC} and BAG~\cite{Cao2019BAGBA}, a gating mechanism is applied on the update vector $u_i^l$ and the hidden state of node $i$ for updating the hidden state of node $i$ at the next layer.
\begin{eqnarray}
w_i^l &=& \sigma(f_g([u_i^l;h_i^l])) \label{eq:1}\\
h_i^{l+1} &=& w_i^l \odot \tanh(u_i^l)+(1-w_i^l) \odot h_i^l  \label{eq:2}
\end{eqnarray}
where $\sigma$ is the sigmoid function, $[u_i^l;h_i^l]$ is the concatenation of $u_i^l$ and $h_i^l$, $f_g$ is implemented with a single-layer multi-layer perceptron (MLP), $\tanh(\cdot)$ is a non-linear activation function, and $\odot$ denotes element-wise multiplication.

This gating mechanism regulates how much of the update message propagates to the next step, so it can prevent overwriting past information. However, the traditional GCN method only updates the central node based on the aggregated information of adjacent nodes, but there is a lot of irrelevant information during information propagating.

When humans do reasoning problems, they always choose the supports information based on the query information. Inspired by human reasoning processing,  we add another question-aware gate to optimize graph convolution procedure, which is suitable for multi-hop reading comprehension. This gating mechanism can regulate the aggregated message according to the question, and introduce the question information into the update message simultaneously.

First, we represent the question $q$ using a bidirectional LSTM (BiLSTM) network~\cite{Hochreiter1997LongSM}, where GLoVe is used as word embeddings.
$$
p = BiLSTM(q), p\in \mathbb{R}^{M \times d}
$$
Then, the final question representation can be obtained by a weighted sum of these vectors.
\begin{eqnarray}
\nonumber w_{ij} &=& \sigma(W_q^T[u_i^l;p_j]+b_q) \\
\nonumber  \alpha_{ij} &=& \frac{\exp(w_{ij})}{\sum_{k=1}^{M}\exp(w_{ik})} \\
\nonumber q_i^l &=& \sum_{j=1}^{M} \alpha_{ij} p_j
\end{eqnarray}
Finally, $u_i^l$ could be updated via:
\begin{eqnarray}
\nonumber  \beta_i^l &=& \sigma(W_s^T[q_i^l;u_i^l] + b_s) \\
\nonumber  u_i^l &=&  \beta_i^l \odot \tanh(q_i^l) +  (1-\beta_i^l) \odot u_i^l
\end{eqnarray}
With this new $u_i^l$, we use Equation~\ref{eq:1} and \ref{eq:2} to obtain $h_i^{l+1}$, which is the hidden state of node $i$ at the $(l+1)$th layer.

We stack the networks for $L$ layers where all parameters are shared, and finally obtain $h^L=\{h^L_i\}_{i=1}^T$ for the entity graph.
\subsection{Output Layer}
Similar to BAG~\cite{Cao2019BAGBA}, we apply a bi-directional attention between the entity graph and the question. The similarity matrix $S\in \mathbb{R}^{T\times M}$ is first calculated via
$$
S=avg_{-1} f_a([h^L;p;h^L \odot p])
$$
where $avg_{-1}$ is the average operation in last dimension, and $f_a$ is a single-layer MLP. Then, the node-to-question attention $g_{n2q}$ and question-to-node attention $g_{q2n}$ are calculated via
\begin{eqnarray}
\nonumber  g_{n2q} &=& softmax_{col}(S)\cdot p \\
\nonumber  g_{q2n} &=& dup(softmax(max_{col}(S)))^T \cdot h^L
\end{eqnarray}
where $softmax_{col}$ and $max_{col}$ denote performing softmax and max function across columns respectively, $dup$ is the function to duplicate the result $softmax(max_{col}(S)) \in \mathbb{R}^{1\times M}$ for $T$ times into shape $\mathbb{R}^{T \times M}$.

The output of the bi-directional attention layer is $[h^L;g_{n2q};h^L\odot g_{n2q};h^L\odot g_{q2n}]$, which is then fed to a 2-layer fully connected feed-forward network with $\tanh$ as the activation function in each layer. Finally, the softmax function is applied among the output, which can generate the prediction result for each node in the graph. Each candidate may correspond to several nodes, since it may appear in multiple documents. We use the maximal probability of these nodes as the result of the candidate, and use cross-entropy as the loss function.

\section{Experiments}
\subsection{Dataset and Experimental Settings}
We use \textsc{WikiHop}~\cite{Welbl2018ConstructingDF} to validate the effectiveness of our proposed approach, which is a multi-choice style reading comprehension data set. The dataset contains about 43K/5K/2.5K samples in training, development, and test set respectively. The test set is not public and can only be evaluated online blindly.

In our implementation, we use NLTK~\cite{bird2006nltk} to tokenize the supporting documents, question, and candidates into word sequences, and then find mentions of subject entity and candidates in supporting documents through the exact matching strategy. In order to extract reasoning entities, we use Stanford CoreNLP~\cite{Manning2014TheSC} to perform entity recognition on the supporting documents.

We use the standard 1024-dimension ELMo and 300-dimension pre-trained GLoVe (trained from 840B Web crawl data) as word embeddings. The dimensions of hidden states in BiLSTM and GCN are set as $d=256$, and the number of nodes and the query length is truncated as 600 and 25 respectively. We stack $L=4$ layers of the Gated-RGCN blocks. During training, we set the mini-batch size as 16, and use Adam~\cite{Kingma2014AdamAM} with learning rate 0.0002 for learning the parameters.

\subsection{Main Results}
We compare our approach with several previously published models, and present our results in Table~\ref{tab:results}. The performance of multiple choice QA is evaluated by the accuracy of choosing the correct answer. Table~\ref{tab:results} shows the performance of approaches both on development and test set respectively. As shown in the table, we can see that our approach achieves the state-of-the-art accuracy on both development and test set against all types of approaches, including GCN-based approaches (Entity-GCN~\cite{de2019question}, BAG~\cite{Cao2019BAGBA}, HDE~\cite{Tu2019MultihopRC}, MHQA-GRN~\cite{Song2018ExploringGP}), memory-based approaches (Coref-GRU~\cite{Dhingra2018NeuralMF}, EPAr~\cite{Jiang2019ExplorePA}), path-based approaches (PathNet~\cite{DBLP:conf/acl/KunduKSC19}), and attention-based approaches (CFC~\cite{Zhong2019CoarsegrainFC}, DynSAN~\cite{Zhuang2019TokenlevelDS}).

For ensemble models, our approach also achieves state-of-the-art performance, which surpasses the reported human performance~\cite{Welbl2018ConstructingDF} by about $4.2\%$.

\begin{table}[htb]
	\centering
	\begin{tabular}{lcc}
		\toprule
		\textbf{Single Models} & \textbf{Dev} & \textbf{Test} \\
		\midrule
		BIDAF~\cite{Seo2016BidirectionalAF}  & 49.7 & 42.9 \\
		Coref-GRU~\cite{Dhingra2018NeuralMF} & 56.0& 59.3 \\		
		MHQA-GRN~\cite{Song2018ExploringGP} & 62.5& 65.4 \\
		Entity-GCN~\cite{de2019question} & 64.8& 67.6 \\
		PathNet$^\dag$~\cite{DBLP:conf/acl/KunduKSC19} & 67.1 & - \\
		BAG~\cite{Cao2019BAGBA} & 66.5 & 69.0 \\
		EPAr~\cite{Jiang2019ExplorePA} & 67.2 & 69.1  \\
		CFC~\cite{Zhong2019CoarsegrainFC} &66.4 &70.6 \\
		HDE~\cite{Tu2019MultihopRC} &68.1 &70.9 \\
		DynSAN~\cite{Zhuang2019TokenlevelDS} & 70.1 &71.4 \\
		\midrule
		\textbf{Proposed} & {\bf70.8} & {\bf72.5} \\
		\bottomrule
		\toprule
		\textbf{Ensemble Models} &  & \\
		\midrule
		Entity-GCN~\cite{de2019question} &68.5 & 71.2 \\
		DynSAN$^\ddag$~\cite{Zhuang2019TokenlevelDS} & - & 73.8 \\
		HDE$^\ddag$~\cite{Tu2019MultihopRC} & - & 74.3 \\
		\midrule
		\textbf{Proposed} & \textbf{74.0} & \textbf{78.3} \\
		\toprule
		Human & - &74.1 \\
		\bottomrule
	\end{tabular}
	\caption{The performance of different models on \textsc{WikiHop} dataset. "-" indicates missing results, $\dag$ indicates that the missing results are not reported in their papers, and $\ddag$ indicates that the results are not reported in their papers but available on \textsc{WikiHop} leaderboard.
	}\label{tab:results}
\end{table}
\begin{figure*}[htp]
	\center
	\subfigure[number of hops]{
		\label{fig:ana:hop}
		\includegraphics[width=0.33\linewidth]{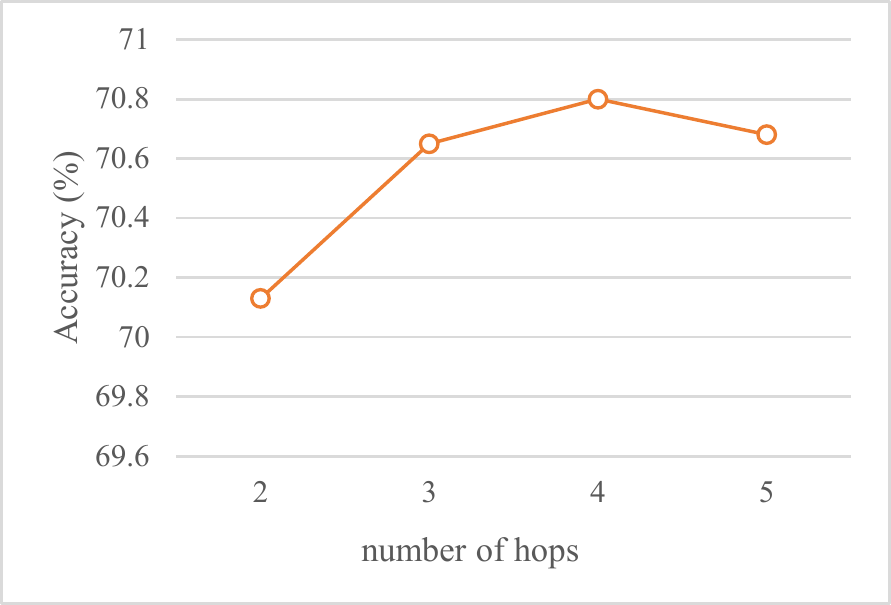}}
	\subfigure[number of supporting docs]{
		\label{fig:ana:sup}
		\includegraphics[width=0.33\linewidth]{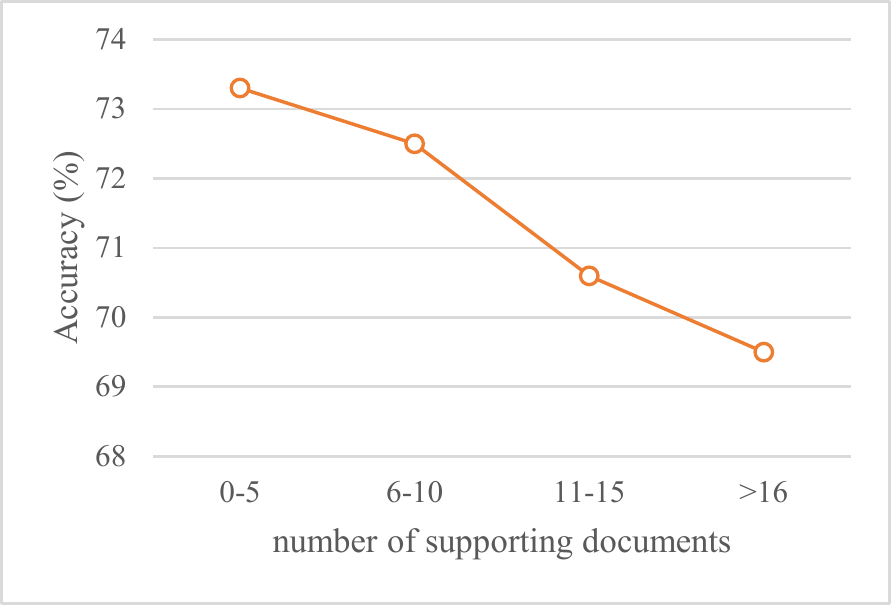}}
	\caption{The results of our approach on the \textsc{WikiHop} dev dataset with different experimental settings.}
	\label{fig:ana}
\end{figure*}
\subsection{Ablations Studies}
We conduct an ablation study to evaluate the contribution of each model component, and show the results in Table~\ref{tab:ablation}.

\begin{table}[htb]
	\centering
	\begin{tabular}{lcc}
		\toprule
		\textbf{Model} & \textbf{Dev} & \textbf{$\triangle$} \\
		\midrule
		\textbf{Full model} & \textbf{70.8} & - \\
		\midrule
		(a) w/o reasoning entities & 69.1 & -1.7 \\
		(b) w/o question &70.2 &-0.6 \\
		(c) w/o attention in question encoding &68.8 &-2.0 \\
		(d) w/o edge types &69.0 &-1.8 \\
		(e) reduce edge types &69.5 &-1.3 \\
		\bottomrule
	\end{tabular}
	\caption{Ablation results on the \textsc{WikiHop} dev set.}\label{tab:ablation}
\end{table}

In (a), we delete all reasoning entities, so our graph degenerates to the entity graph in ~\cite{de2019question}. The accuracy on the development set drops to 69.1\%, but it is still higher than the accuracy of Entity-GCN~\cite{de2019question}. This proves the effectiveness of reasoning entities and the Gated-RGCN mechanism. In (b), we do not bring in the question in the reasoning. While in (c), we only use BiLSTM to encode the question, but do not adopt the attention mechanism in BiLSTM. From (b) and (c), we can see that question is useful for reasoning. In (d), we treat all edge types equally, so Gated-RGCN is replaced by Gated-GCN in reasoning, which reduces $1.8\%$ accuracy absolutely. In (e), we only define the edge types as in BAG~\cite{Cao2019BAGBA}, and the accuracy drops to $69.5\%$. From (d) and (e), we learn that different types of edges is also critical in the reasoning.

\subsection{Analysis}
In this section, we conduct a series of experiments with different setting in our approach.

\paragraph{Different number of Gated-RGCN layers.} We evaluate our approach with different number of GCN layers ($L$), and the result is shown in Figure~\ref{fig:ana:hop}. From the figure, we can see that the accuracy increases gradually and then drops finally, which proves the effectivity of Gated-RGCN. It is expected that the accuracy drops finally, because more hops would bring in noise. As for why 4-hops is the best, this may be because many samples are 1-hop problems, and the length of path in these samples is exactly 4.

\paragraph{Different number of supporting documents.} We split the development set into subsets according to the number of supporting documents, and then evaluate our approach on each subset. The results are shown in Figure~\ref{fig:ana:sup}. From the figure, we can see that more documents may bring in more irrelevant information, which is harmful to the multi-hop QA. However, even on a number greater than 16, our model achieved an accuracy of 69.5\%, which is better than the overall effect of most models.

\paragraph{Different word embeddings.} Table~\ref{tab:embed} shows the results of our approach when using different word embeddings. From the table, we can see that only using ELMo or GLoVe in our approach will cause a severe drop, and replacing ELMo with BERT~\cite{Devlin2019BERTPO} delivers a competitive result. Meanwhile, we can see that for the GCN-based approach, the initialization of nodes is extremely critical, because this type of approach has greatly compressed the information when constructing the graph.

\begin{table}[htp]
	\centering
	\begin{tabular}{lc}
		\toprule
		\textbf{Embeddings} & \textbf{Dev}  \\
		\midrule
		ELMo & 66.1  \\
		GLoVe &64.5 \\
		Bert+GLoVe &70.78 \\
		ELMo+GLoVe &\textbf{70.81} \\
		\bottomrule
	\end{tabular}
	\caption{The results with different word embeddings on the \textsc{WikiHop} dev set.}\label{tab:embed}
\end{table}

\section{Conclusion}

In this paper, we propose a novel approach for multi-hop reading comprehension across documents. Our approach extends the entity graph by introducing reasoning entities, which can form the reasoning path from question to candidates. In addition, our approach incorporates the question in the multi-hop reasoning through a new gate mechanism to regulate how much useful information propagating from neighbors to the node. Experiments show that our approach achieves state-of-the-art accuracy both for single and ensemble models.

Our future work would focus on the interpretability of the multi-hop reading comprehension across documents. In addition, we would like to build the entity graph with reasoning entities dynamically during the reasoning, and apply our model on other datasets.

\section*{Acknowledgments}
This work is supported by the National Key Research and Development Project of China (No. 2018AAA0101900), the Fundamental Research Funds for the Central Universities (No. 2019FZA5013), the Zhejiang Provincial Natural Science Foundation of China (No. LY17F020015), the Chinese Knowledge Center of Engineering Science and Technology (CKCEST) and MOE Engineering Research Center of Digital Library.

\bibliographystyle{named}
\bibliography{ijcai20}

\begin{thebibliography}{}

\bibitem[\protect\citeauthoryear{Bird}{2006}]{bird2006nltk}
Steven Bird.
\newblock Nltk: the natural language toolkit.
\newblock In {\em Proceedings of the COLING/ACL on Interactive presentation
  sessions}, pages 69--72. Association for Computational Linguistics, 2006.

\bibitem[\protect\citeauthoryear{Cao \bgroup \em et al.\egroup
  }{2019}]{Cao2019BAGBA}
Yu~Cao, Meng Fang, and Dacheng Tao.
\newblock Bag: Bi-directional attention entity graph convolutional network for
  multi-hop reasoning question answering.
\newblock In {\em NAACL-HLT}, 2019.

\bibitem[\protect\citeauthoryear{Das \bgroup \em et al.\egroup
  }{2017}]{Das2017GoFA}
Rajarshi Das, Shehzaad Dhuliawala, Manzil Zaheer, Luke Vilnis, Ishan Durugkar,
  Akshay Krishnamurthy, Alexander~J. Smola, and Andrew McCallum.
\newblock Go for a walk and arrive at the answer: Reasoning over paths in
  knowledge bases using reinforcement learning.
\newblock In {\em ICLR}, 2017.

\bibitem[\protect\citeauthoryear{De~Cao \bgroup \em et al.\egroup
  }{2019}]{de2019question}
Nicola De~Cao, Wilker Aziz, and Ivan Titov.
\newblock Question answering by reasoning across documents with graph
  convolutional networks.
\newblock In {\em Proceedings of the 2019 Conference of the North American
  Chapter of the Association for Computational Linguistics: Human Language
  Technologies, Volume 1 (Long and Short Papers)}, pages 2306--2317, 2019.

\bibitem[\protect\citeauthoryear{Devlin \bgroup \em et al.\egroup
  }{2019}]{Devlin2019BERTPO}
Jacob Devlin, Ming-Wei Chang, Kenton Lee, and Kristina Toutanova.
\newblock Bert: Pre-training of deep bidirectional transformers for language
  understanding.
\newblock In {\em NAACL-HLT}, 2019.

\bibitem[\protect\citeauthoryear{Dhingra \bgroup \em et al.\egroup
  }{2018}]{Dhingra2018NeuralMF}
Bhuwan Dhingra, Qiao Jin, Zhilin Yang, William~W. Cohen, and Ruslan
  Salakhutdinov.
\newblock Neural models for reasoning over multiple mentions using coreference.
\newblock In {\em NAACL-HLT}, 2018.

\bibitem[\protect\citeauthoryear{Hochreiter and
  Schmidhuber}{1997}]{Hochreiter1997LongSM}
Sepp Hochreiter and J{\"u}rgen Schmidhuber.
\newblock Long short-term memory.
\newblock {\em Neural Computation}, 9:1735--1780, 1997.

\bibitem[\protect\citeauthoryear{Hu \bgroup \em et al.\egroup
  }{2017}]{Hu2017ReinforcedMR}
Minghao Hu, Yuxing Peng, Zhen Huang, Xipeng Qiu, Furu Wei, and Ming Zhou.
\newblock Reinforced mnemonic reader for machine reading comprehension.
\newblock In {\em IJCAI}, 2017.

\bibitem[\protect\citeauthoryear{Jiang \bgroup \em et al.\egroup
  }{2019}]{Jiang2019ExplorePA}
Yichen Jiang, N.~Joshi, Yen-Chun Chen, and Mohit Bansal.
\newblock Explore, propose, and assemble: An interpretable model for multi-hop
  reading comprehension.
\newblock In {\em ACL}, 2019.

\bibitem[\protect\citeauthoryear{Khashabi \bgroup \em et al.\egroup
  }{2018}]{Khashabi2018LookingBT}
Daniel Khashabi, Snigdha Chaturvedi, Michael Roth, Shyam Upadhyay, and Dan
  Roth.
\newblock Looking beyond the surface: A challenge set for reading comprehension
  over multiple sentences.
\newblock In {\em NAACL-HLT}, 2018.

\bibitem[\protect\citeauthoryear{Kingma and Ba}{2015}]{Kingma2014AdamAM}
Diederik~P. Kingma and Jimmy Ba.
\newblock Adam: A method for stochastic optimization.
\newblock In {\em ICLR}, 2015.

\bibitem[\protect\citeauthoryear{Kocisk{\'y} \bgroup \em et al.\egroup
  }{2017}]{Kocisk2017TheNR}
Tom{\'a}s Kocisk{\'y}, Jonathan Schwarz, Phil Blunsom, Chris Dyer, Karl~Moritz
  Hermann, G{\'a}bor Melis, and Edward Grefenstette.
\newblock The narrativeqa reading comprehension challenge.
\newblock {\em Transactions of the Association for Computational Linguistics},
  6:317--328, 2017.

\bibitem[\protect\citeauthoryear{Kundu \bgroup \em et al.\egroup
  }{2019}]{DBLP:conf/acl/KunduKSC19}
Souvik Kundu, Tushar Khot, Ashish Sabharwal, and Peter Clark.
\newblock Exploiting explicit paths for multi-hop reading comprehension.
\newblock In {\em {ACL} {(1)}}, pages 2737--2747. Association for Computational
  Linguistics, 2019.

\bibitem[\protect\citeauthoryear{Lv \bgroup \em et al.\egroup
  }{2019}]{Lv2019AdaptingMK}
Xin Lv, Yuxian Gu, Xu~Han, Lei Hou, Juanzi Li, and Zhiyuan Liu.
\newblock Adapting meta knowledge graph information for multi-hop reasoning
  over few-shot relations.
\newblock {\em ArXiv}, abs/1908.11513, 2019.

\bibitem[\protect\citeauthoryear{Manning \bgroup \em et al.\egroup
  }{2014}]{Manning2014TheSC}
Christopher~D. Manning, Mihai Surdeanu, John Bauer, Jenny~Rose Finkel, Steven
  Bethard, and David McClosky.
\newblock The stanford corenlp natural language processing toolkit.
\newblock In {\em ACL}, 2014.

\bibitem[\protect\citeauthoryear{Mihaylov \bgroup \em et al.\egroup
  }{2018}]{Mihaylov2018CanAS}
Todor Mihaylov, Peter Clark, Tushar Khot, and Ashish Sabharwal.
\newblock Can a suit of armor conduct electricity? a new dataset for open book
  question answering.
\newblock In {\em EMNLP}, 2018.

\bibitem[\protect\citeauthoryear{Pennington \bgroup \em et al.\egroup
  }{2014}]{Pennington2014GloveGV}
Jeffrey Pennington, Richard Socher, and Christopher~D. Manning.
\newblock Glove: Global vectors for word representation.
\newblock In {\em EMNLP}, 2014.

\bibitem[\protect\citeauthoryear{Peters \bgroup \em et al.\egroup
  }{2018}]{peters2018deep}
Matthew~E Peters, Mark Neumann, Mohit Iyyer, Matt Gardner, Christopher Clark,
  Kenton Lee, and Luke Zettlemoyer.
\newblock Deep contextualized word representations.
\newblock In {\em Proceedings of NAACL-HLT}, pages 2227--2237, 2018.

\bibitem[\protect\citeauthoryear{Qiu \bgroup \em et al.\egroup
  }{2019}]{Qiu2019DynamicallyFG}
Lin Qiu, Yunxuan Xiao, Yanru Qu, Hao Zhou, Lei Li, Weinan Zhang, and Yong Yu.
\newblock Dynamically fused graph network for multi-hop reasoning.
\newblock In {\em ACL}, 2019.

\bibitem[\protect\citeauthoryear{Seo \bgroup \em et al.\egroup
  }{2017}]{Seo2016BidirectionalAF}
Minjoon Seo, Aniruddha Kembhavi, Ali Farhadi, and Hannaneh Hajishirzi.
\newblock Bidirectional attention flow for machine comprehension.
\newblock In {\em ICLR}, 2017.

\bibitem[\protect\citeauthoryear{Shen \bgroup \em et al.\egroup
  }{2017}]{shen2017reasonet}
Yelong Shen, Po-Sen Huang, Jianfeng Gao, and Weizhu Chen.
\newblock Reasonet: Learning to stop reading in machine comprehension.
\newblock In {\em Proceedings of the 23rd ACM SIGKDD International Conference
  on Knowledge Discovery and Data Mining}, pages 1047--1055. ACM, 2017.

\bibitem[\protect\citeauthoryear{Song \bgroup \em et al.\egroup
  }{2018}]{Song2018ExploringGP}
Linfeng Song, Zhiguo Wang, Mo~Yu, Yue Zhang, Radu Florian, and Daniel Gildea.
\newblock Exploring graph-structured passage representation for multi-hop
  reading comprehension with graph neural networks.
\newblock {\em ArXiv}, abs/1809.02040, 2018.

\bibitem[\protect\citeauthoryear{Tu \bgroup \em et al.\egroup
  }{2019}]{Tu2019MultihopRC}
Ming Tu, Guangtao Wang, Jing Huang, Yun~Shu Tang, Xiaodong He, and Bowen Zhou.
\newblock Multi-hop reading comprehension across multiple documents by
  reasoning over heterogeneous graphs.
\newblock In {\em ACL}, 2019.

\bibitem[\protect\citeauthoryear{Welbl \bgroup \em et al.\egroup
  }{2018}]{Welbl2018ConstructingDF}
Johannes Welbl, Pontus Stenetorp, and Sebastian Riedel.
\newblock Constructing datasets for multi-hop reading comprehension across
  documents.
\newblock {\em Transactions of the Association for Computational Linguistics},
  6:287--302, 2018.

\bibitem[\protect\citeauthoryear{Williams}{1992}]{Williams1992SimpleSG}
Ronald~J. Williams.
\newblock Simple statistical gradient-following algorithms for connectionist
  reinforcement learning.
\newblock {\em Machine Learning}, 8:229--256, 1992.

\bibitem[\protect\citeauthoryear{Yang \bgroup \em et al.\egroup
  }{2018}]{Yang2018HotpotQAAD}
Zhilin Yang, Peng Qi, Saizheng Zhang, Yoshua Bengio, William~W. Cohen, Ruslan
  Salakhutdinov, and Christopher~D. Manning.
\newblock Hotpotqa: A dataset for diverse, explainable multi-hop question
  answering.
\newblock In {\em EMNLP}, 2018.

\bibitem[\protect\citeauthoryear{Zhong \bgroup \em et al.\egroup
  }{2019}]{Zhong2019CoarsegrainFC}
Victor Zhong, Caiming Xiong, Nitish~Shirish Keskar, and Richard Socher.
\newblock Coarse-grain fine-grain coattention network for multi-evidence
  question answering.
\newblock In {\em ICLR}, 2019.

\bibitem[\protect\citeauthoryear{Zhuang and
  Wang}{2019}]{Zhuang2019TokenlevelDS}
Yimeng Zhuang and Huadong Wang.
\newblock Token-level dynamic self-attention network for multi-passage reading
  comprehension.
\newblock In {\em ACL}, 2019.

\end{thebibliography}

\end{document}